# *AI's Spatial Intelligence: Evaluating AI's Understanding of Spatial Transformations in PSVT:R and Augmented Reality*

Uttamasha Monjoree and Wei Yan

Texas A&M University

**Abstract**

Spatial intelligence is important in many fields such as Architecture, Engineering, and Construction (AEC), Science, Technology, Engineering, and Mathematics (STEM), and Medicine. Understanding three-dimensional (3D) spatial rotations can involve verbal descriptions and visual or interactive examples, illustrating how objects move and change orientation in 3D space. Recent studies show Artificial Intelligence (AI) with language and vision capabilities still face limitations in spatial reasoning. In this paper, we have studied advanced generative AI's spatial capabilities of understanding rotations of objects in 3D space utilizing its image processing and language processing features. We examined the spatial intelligence of three AI vision models (GPT-4, Gemini 1.5 Pro, and Llama 3.2) in understanding spatial rotation process with spatial rotation diagrams based on the Revised Purdue Spatial Visualization Test: Visualization of Rotations (Revised PSVT:R). Furthermore, we incorporated an added layer of a coordinate system axes on Revised PSVT:R to study the variations in AI vision models' performance. We additionally examined AI vision models' understanding of 3D rotations in Augmented Reality (AR) scene images that visualize spatial rotations of a physical object in 3D space and observed increased accuracy of AI vision models' understanding of the rotations by adding supplementary textual information depicting the rotation process or mathematical representations of the rotation (e.g., matrices) superimposed on the object. The results indicate that while GPT-4, Gemini 1.5 Pro, and Llama 3.2 as major current Generative AI model lacks the understanding of a spatial rotation process, it has the potential to understand the rotation process with additional information that can be provided by methods such as AR. AR can superimpose textual information or mathematical representations of the rotations on spatial transformation diagrams or images and create a more intelligible input for AI to comprehend or for training AI's spatial intelligence. Furthermore, by combining the potentials in spatial intelligence of AI with AR's interactive visualization abilities, we expect to offer enhanced guidance for students' spatial learning activities. Such spatial guidance can greatly benefit understanding spatial transformations and additionally support processes like assembly, construction, fabrication, manufacturing, as well as learning in AEC, STEM, and Medicine that require precise 3D spatial understanding and instruction.

# 1. Introduction

Spatial visualization skills refer to the ability to mentally rotate, manipulate, twist, or invert 3D objects [1], [2]. Education in many STEM disciplines demands instructional approaches that offer experimental practices and tasks with high representational accuracy and realistic simulations to enhance the learning experience [3]. Many first-year undergraduate STEM students struggle with visualization and drawing tasks, even though they have access to various computer assisted learning resources for geometry, spatial transformations, and related mathematical concepts [4]. These difficulties include understanding the symbolic notation of linear algebra, generalizing geometric reasoning, and transitioning to matrix representations of transformations, which often weakens their intuitive grasp of the concepts [5], [6].

Despite studying drawing in secondary school, many students have not sufficiently developed their spatial abilities, leading to significant challenges in mentally manipulating figures in space [7].

Personalized STEM education using interactive technologies can enhance learning experiences. It can promote engagement through hands-on simulations and immersive environments and provide real-time feedback to improve understanding of complex concepts [3]. By leveraging tools such as augmented reality (AR), virtual labs, and Artificial Intelligence (AI)-driven assistance, students can explore STEM topics at their own pace and develop critical problem-solving skills in an interactive, engaging way [8].

Generative AI, encompassing technologies that can create new content such as text, images, 3D models, and audio, have rapidly evolved in recent years, driven by advancements in deep learning and neural network architecture. Gemini Pro is capable of a variety of NLP tasks including content understanding, text summarization, object recognition, classification and extrapolation as a foundational model [9], [10]. Llama is specifically designed to support image reasoning use cases and serves as a foundation to build more robust content moderation tools for human-AI conversation with multimodal capabilities [11], [12]. Large language models (LLMs) such as GPT-4 and its successors have demonstrated remarkable proficiency in generating coherent and contextually relevant text. GPT-4 with Vision - GPT-4V(ision) - has capabilities to analyze user-input images and show unprecedented ability in understanding visual inputs, spatial relationships of humans and objects, and the genericity of its capabilities [13],[14], [15].

Augmented Reality (AR) can overlay digital information on the real world and create a more perceivable representation for AI. The use of AR-based technologies can reduce the cognitive load of users required for spatial learning by superimposing virtual instruction on the real world. AR has three distinguishing features: combining real and virtual; interactive in real-time; and registered in three dimensions [16]. Using AR in education has several positive impacts: increased content understanding, learning spatial structures, language associations, long-term memory retention, improved collaboration, and motivation [17].

Generative AI's visual capabilities are highly relevant in AR, virtual design, and other applications that require real-time understanding of visual input [18]. For instance, it can assist in designing objects in 3D space [19], guide users through augmented experiences, or help in manipulating visual content interactively [20]. Combined textual and graphical information superimposed in real world utilizing AR technology has the potential to generate a more intelligible input for AI vision models' capabilities.

Using conversational AI in education is one of the major approaches to enhancing and promoting a more personalized learning experience [21]. Personalized interactive assistance of generative conversational AI that has visual capabilities along with spatial qualities of AR can help students achieve better learning outcomes. Understanding AI's ability to perceive rotation in AR is expected to have a significant impact on learning spatial visualization skills.

Spatial reasoning is a key area in psychometrics, and several datasets exist for evaluating these skills, such as the Mental Rotation Tests [22], the Purdue Spatial Visualization Test (PSVT)/The Purdue Visualization of Rotations Test [23], and the Revised Purdue Spatial Visualization Test: Visualization of Rotations (Revised PSVT:R) [24]. The Revised PSVT:R is considered one of the most effective tools for evaluating spatial reasoning skills. This test is a widely used assessment designed to measure an individual's ability to visualize and mentally manipulate 3D objects in space, particularly in terms of rotating these objects.

We can use both a verbal description and some visual or interactive examples to explain spatial rotations and convey how objects move or change orientations in 3D space [25]. Combining the visual and verbal abilities of conversational AI and visual and interactive abilities of AR has the potential to provide enhanced guidance in spatial rotation activities [8]. This type of spatial guidance can help mathematical education as well as govern assembly, construction, and fabrication process that requires spatial instructions. The goal of this study is to evaluate AI's capability of understanding 3D spatial rotations by performing a series of tests based on Revised PSVT:R and an interactive AR application (AR-Classroom) that can visualize rotation along different axes [26]. The research questions that guide this study are as follows:

1. Can GPT-4 answer questions in Revised PSVT:R and how does its performance compare with human test takers?
2. Can GPT-4 perceive a spatial rotation process created based on Revised PSVT:R and describe the process in terms of axis of rotation, angle of rotation, and (clockwise or counterclockwise) direction of rotation?
3. Can GPT-4 perceive a spatial rotation process visualized in AR and describe the process in terms of axis of rotation, angle of rotation, and (clockwise or counterclockwise) direction of rotation?

## 2. Literature Review

*Dual-coding theory*

The dual-coding theory is defined as a theory that posits the existence of two interconnected subsystems in the human cognitive system: a verbal system and an imagery system [27]. This dual capacity means the language system functions in both communication and symbolic interpretation. Human cognition uniquely integrates both language and nonverbal objects and events, enabling us to process linguistic input (like speech or writing) alongside symbolic representations of nonverbal aspects of the world [28]. Therefore, spatial representation should integrate these two systems, where language serves both direct communication and a framework for understanding nonverbal phenomena. Integrating sophisticated language and image processing and response generation of Artificial Intelligence (AI) with immersive three-dimensional capabilities of AR can facilitate a more engaging spatial learning experience [29]. Including Augmented Reality (AR) in the learning process can improve spatial abilities in the domain of architecture education [30].

*AI Vision Model Application*

The latest development of large-scale, multimodal AI model can accept image and text inputs and produce text outputs. While less capable than humans in many real-world scenarios, GPT-4 Vision, Gemini 1.5 Pro, Llama 3.2 can analyze images, provide detailed descriptions, identify objects, and offer explanations of scenes [9], [10], [14]. AI with vision and spatial understanding have broad applications, for example, in STEM and Architecture/Engineering/Construction (AEC). Integrating GPT AI models into construction safety management enhances risk assessment, real-time insights, incident reporting, and compliance evaluation [31], e.g. the Personal Protective Equipment (PPE) or helmet usage compliance [15]. Experiments have proven GPT-4V to be highly effective in detecting and explaining global and fine-grained semantic patterns in anomaly or risk detection [32], [15].

*AI's Spatial and Mathematical Abilities*

Spatial intelligence is described as a skillset that helps us comprehend visual-spatial tasks and spatial relations. As a result, we gain a better orientation of objects in space [33]. In recent years there have been several studies analyzing spatial and mathematical capabilities of AI. GPT-4 model "hallucinates", wherein the model returned inaccurate information. The model was unable to accurately return bounding boxes for object detection, suggesting it is not completely fit for computer vision related usage [34]. The SPARE3D dataset introduces a series of challenging 2D-3D spatial reasoning tasks, revealing that while convolutional neural networks (CNNs) excel in visual learning, their performance on spatial reasoning remains below human average, highlighting the need for advancements in artificial neural network design to improve 3D spatial reasoning for intelligent robots [35]. A recent study has examined the perceived complexity of 3D shapes from both human and ChatGPT perspectives, revealing that ChatGPT can effectively capture human consensus on shape complexity. The findings suggest that large generative models like ChatGPT could be used to automatically generate 3D content with varying complexities, which can be enhanced for spatial visualization skills in VR applications [18]. This research explored how the visual features of some of the 3D shapes used in the Purdue Spatial Visualization Test are perceived by ChatGPT.

After an elaborate study, [36] observed state of the art LLMs ( Llama3-8b, GPT-4o, Gemma2-9B, Phi-3, Phi-3.5, and Mathstral-7B) resemble sophisticated pattern matching more than true logical reasoning. LLM performance varies highly on different versions of the same question. LLMs' performance drops substantially with a minor increase in difficulty, and their sensitivity to inconsequential information indicates that their reasoning is fragile. Another study found, ChatGPT-4 and ChatGPT 4o struggle more with geometry and complex math problems compared to algebra, suggesting that while AI can support algebra instruction, teacher intervention is needed in areas where AI is less effective, such as spatial intelligence and higher-difficulty tasks [37].

Despite many positive media reports highlighting LLMs' exam-solving abilities, possibly influenced by a selection bias, their overall mathematical performance falls significantly short of a graduate student's level [38]. Educators can leverage AI to create engaging and effective learning environments, cultivating a generation skilled in mathematics and prepared to navigate the complexities of the modern world by addressing the associated challenges and adopting a refined approach [39].

*AR and AI in Education*

In recent years, AR and AI have demonstrated the potential to reshape education by enhancing interactivity, personalization, and accessibility. The use of AR technology and speech recognition to support productive vocabulary among students has shown user satisfaction when parents and teachers were interviewed [40]. [41] introduced a novel early education platform that combines smart voice commands and AR technology. The platform is embedded in a storage box, making it convenient for use in homes and kindergartens where preschool education is provided. This project enhances the integration of these technologies in educational settings.

Using AR-powered embodied learning for spatial transformations and representation of corresponding matrices has already shown significant improvement in the participants' math scores [42]. The project AR-Classroom uses embodied learning and novel AR features to visualize spatial rotations and their mathematical representations, and the usability tests that were completed using the application showed promising results. The benchmark test and updated usability test done on 24 participants have shown significant improvement in usability scores [32]. The AR-Classroom application integrates AR and AI-based multi-sensory assistance to help undergraduate students better understand spatial transformations and their mathematical representations. A usability study identified four key themes: ease of use, the need for more detailed AI responses, a desire for visual information, and improved understanding, highlighting the potential for further AI-guided visual cues in AR learning environments [8].

## 3. Methodology

When explaining rotation, we use references like direction, angle, and axis verbally and visually. According to the dual-coding theory, two cognitive subsystems, verbal and imagery, work together to process language and nonverbal content. Integrating these systems has the potential to enhance spatial understanding, where language aids in communication and interpreting nonverbal cues.

LLMs rely on descriptive datasets and visual cues, which can enhance spatial descriptions in vision models. Using visual cues to communicate a rotation process could improve interactions between AR and AI systems, providing a more intuitive tool for conveying spatial information, thus facilitating spatial learning. This approach could bridge the gap between text-based AI and spatially oriented AR, making explanations of spatial transformations more accessible and effective. Combined AR and AI have the potential of becoming instrumental in assembly and fabrication, offering real-time guidance, error reduction, and efficiency improvements. In this paper, we examine generative AI's vision and language capacity of understanding and describing spatial rotation using standard test (Revised PSVT:R) vs AR.

Our evaluation focused on the AI's capacity to comprehend key aspects of spatial rotations, including the axis, angle, and direction of each rotation, as well as its ability to process sequences of multiple rotations. We integrated multiple layers of graphical and textual features into the Revised PSVT:R to analyze their impact on the performance of GPT-4, Gemini 1.5 Pro, and Llama 3.2. We also tested their capability in an AR application that visualizes the spatial rotation of a hand-held object in 3D space and tested the accuracy with the addition of supplementary textual information (angle, textual information, rotation matrix equation etc.)

## 4. Experiments and Results

### *Experiment 1: Evaluate GPT-4 Using Revised PSVT:R (original and with added coordinate axes)*

The Revised PSVT:R is a widely used test for assessing spatial visualization skills, consisting of rotation problems. In this test, participants are shown diagrams of 3D objects in both a starting and finishing orientation after an unknown rotation along one or more axes. The task is to identify the correct diagram of a second 3D object that has undergone the same rotations as the first. Participants are given the second 3D object in its starting orientation and must choose the correct finishing orientation from five possible options.

### *Test Using Revised PSVT:R (Test 1a)*

Using ChatGPT 4o (with a vision capability) from June 9, 2024 through Fall 2024, we conducted various tests. In the first test, we provided GPT-4, Gemini 1.5, and Llama 3.2 with all the 30 problems from Revised Purdue Spatial Visualization Test: Visualization of Rotations (Revised PSVT:R) as a standard test to study its ability to analyze input images and spatial rotation capabilities (Figure 1a). After getting the input (a screenshot of the test questions) uploaded, AI vision models responded with answering the questions by choosing from (A, B, C, D, and E options). (For copyright of Revised PSVT:R and keeping the test results confidential for future tests with participants, the 30 questions and results are not revealed in this paper. An example question in the DIRECTIONS section of the test that is allowed to publish is shown in Figure 1.)

### *Test Using Revised PSVT:R with Axes added (Test 1b)*

After getting unsatisfactory results in the first test, we explored more of its spatial analyzing capabilities by adding a coordinate system with x- y- z-axes onto the Revised PSVT:R question images. The motivation behind this is incorporating more visual and textual cues for analyzing images and enhancing communication with AI. We superimposed the coordinate system on 30 question images of Revised PSVT:R where x-, y-, and z-axes are indicated (Figure 1b).

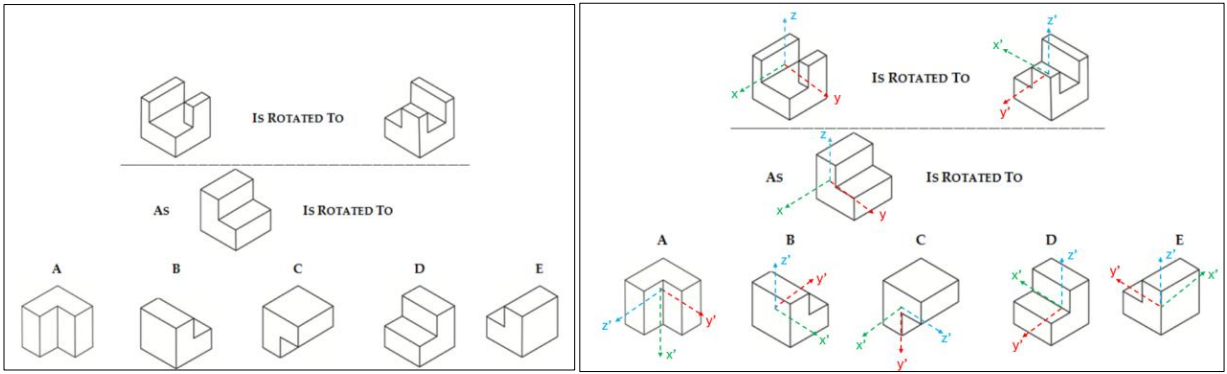

Figure 1: (a) The example question from Revised PSVT:R (left), (b) the example question from Revised PSVT:R with a coordinate system added (right). In both questions, D is the correct answer.

In Experiment 1, both Test 1a and 1b, GPT-4 had to identify the correct option from the five 3D object orientations (A to E). To execute the whole task, like a human test-taker, AI vision models need to identify the rotation process in Step 1 and then compare it to the five rotation processes in Step 2, in order to choose the correct option (Figure 2). For example, in Figure 2, to answer correctly, one should identify the similarities between the axis systems between Step 1 and Step 2, and choose option D. In Test 1a and Test 1b, GPT-4, Gemini 1.5 Pro, and Llama 3.2 gave incorrect answers most of the cases (Figure 3). The accuracy of the LLMs in these two tests varied between 13.3% to 23.3%, which indicates that it is unable to identify and compare rotation processes correctly even with the visual cues (axes) added.

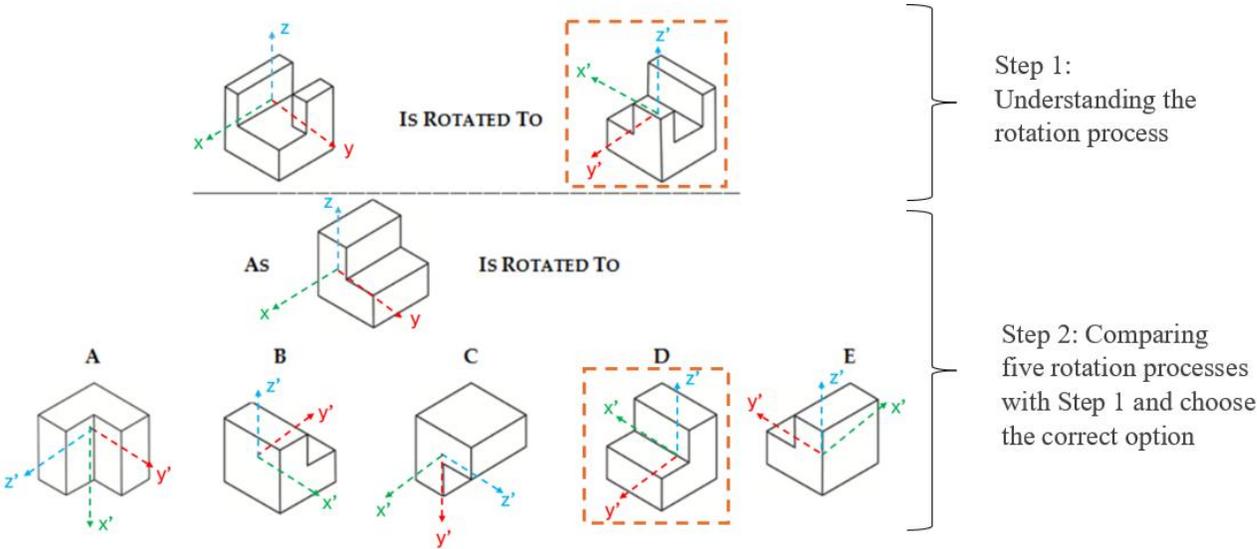

Figure 2: Example question in Revised PSVT:R with a coordinates system. D is the correct answer.

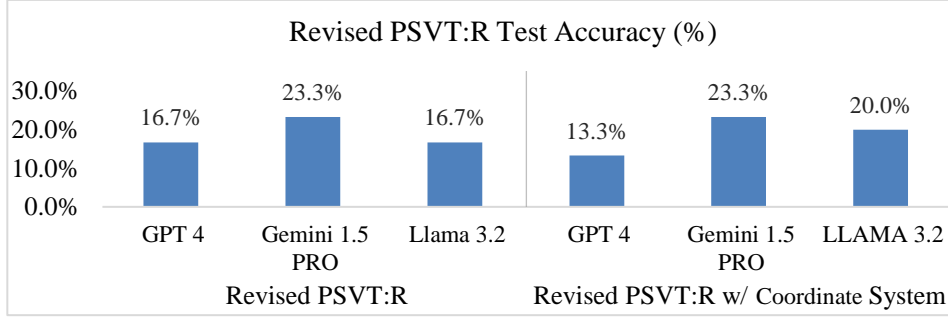

Figure 3: Accuracy of GPT-4, Gemini 1.5 Pro, Llama 3.2 in Experiment 1 using Revised PSVT:R

*Experiment 2: Evaluate GPT-4 Using Step 1 of Revised PSVT:R with a Coordinate System (Test 2)*

GPT-4 showed an unsatisfactory result in answering the questions of Revised PSVT:R in Experiment 1. To simplify the tasks for AI, we continued testing with only Step 1 of the questions of Revised PSVT:R (Figure 3) to see if AI can identify the rotations of one object in each question. We superimposed a coordinate system with axes on Step 1 of each question of Revised PSVT:R, and asked AI vision models to explain the rotation process with minimum rotation steps (Figure 4). The resulting orientation in 14 questions of Revised PSVT:R can be achieved by rotating along a single axis and in the rest 16 questions can be achieved by rotating along multiple axes (e.g., rotating around one axis and then around another axis). As the resulting orientation of a 3D object can be achieved using different sequences of multiple rotations, we asked AI vision models to describe the change of orientation using rotations along a minimum number of axes. We used the prompt, *"Explain the rotation process using minimum steps of rotation in terms of axis of rotation, angle of rotation, and direction of rotation."* From the results we can observe, GPT-4 performed better in Experiment 2 than Gemini 1.5 Pro, and Llama 3.2 (Figure 5).

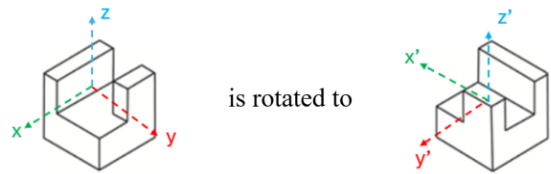

Figure 4: Diagram of Step 1 of Revised PSVT:R with a coordinate system added.

In this test, GPT-4 performed better in case of rotation along a single axis than rotations along multiple axes (Figure 5). In case of a rotation along a single axis we observed 35.7% accuracy, in case of rotations along multiple axes, we observed 12.5% accuracy. The combined accuracy of this test is 23.3%. The performance in case of rotation along a single axis is almost 2 times more accurate than the rotations along multiple axes, though in both cases performance by GPT-4 is unsatisfactory. The accuracy of Gemini 1.5 Pro and Llama 3.2 were both unsatisfactory though Gemini performed slightly better than Llama in case of single axis rotation and combined result. This experiment results indicate that even though these AI vision models cannot compare similar rotation processes, GPT-4 is somewhat capable of understanding one rotation process. The information we provided is not sufficient for these AI vision models to understand the rotation process. However, there may still be a possibility of achieving better results by adding supplementary information and context.

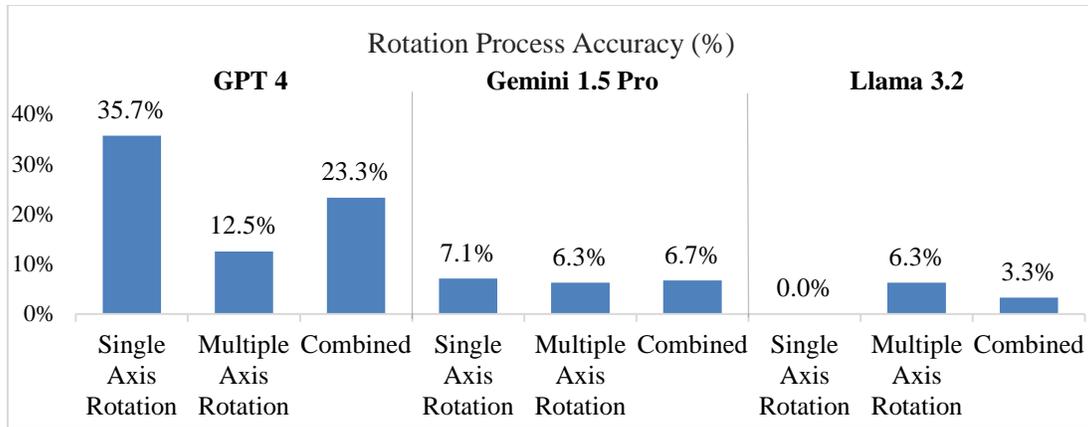

Figure 5: Rotation process accuracy described by GPT-4 using Step 1 of Revised PSVT:R with axes of a coordinate system

*Experiment 3: Evaluate GPT-4 on Understanding Rotations with Supplementary Information through an AR Application*

AR can superimpose 3D models, data, or text over real-world objects, enhancing the user's understanding of their surroundings and the physical world. An AR application named *AR-Classroom* provides an interactive learning environment with better context and visualization for understanding rotations (Figure 6) [26]. The app aims to run on an AR device, such as an AR-enabled tablet, an AR headset, or a computer with a camera. A user can hold and manipulate a 3D physical model to learn spatial transformations by observing the corresponding mathematical representation and visualization that augment the 3D physical model with a graphical user interface (GUI) in AR.

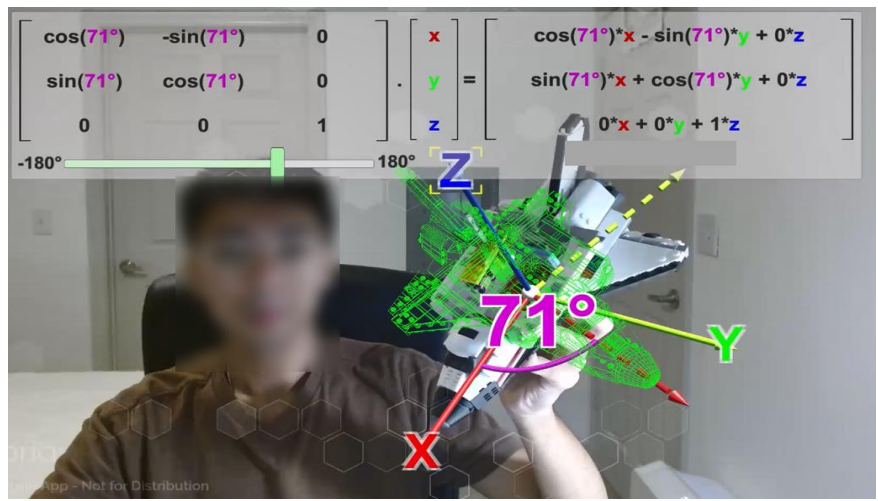

Figure 6: AR-Classroom GUI – augmenting the 3D physical space shuttle model with a 3D virtual space shuttle model, rotation axes, rotation arc and angle, and the matrix representation of the rotation.

The X-, Y-, and Z-axes of the models' coordinate systems and corresponding matrix equations are displayed in the GUI. In Figure 6, the following AR scene components are shown.

1. a physical model, called pre-image of a transformation, showing the state before rotation
2. a 3D reference frame coordinate with red, green, and blue solid arrows for X-, Y-, and Z-axis, attached to the physical model

3. a 3D wireframe model, called image of a transformation, showing the state after rotation
4. a 3D body frame coordinate with red, green, and blue dotted arrows for X-, Y-, and Z-axes, attached to the 3D wireframe model

*Test Using AR with a Coordinate System (Test 3a):* We provided 30 screenshots of the AR environment visualizing only the physical and virtual models and the coordinate system to AI vision models. The angle of rotation and corresponding rotation matrix equation were hidden (Figure 7a). All screenshots varied in axis of rotation, angle of rotation, and direction (clockwise or counterclockwise) of rotation. We provided the prompt, "*Explain the rotation process with axis of rotation, angle of rotation and direction. Note: All red lines represent X-axis, all yellow-green lines represent Y-axis, all blue lines represent Z-axis. The dotted line is the orientation of axis after rotation. The axis of rotation does not have a dotted counterpart of the same color. Follow right hand rule for the direction of rotation*."

*Test Using AR with a Coordinate System and Angle of Rotation (Test 3b):* We provided 30 screenshots of the AR environment visualizing the physical and virtual models, the coordinate system, and angle of rotation to AI vision models. The corresponding rotation matrix equation was hidden (Figure 7b). All screenshots varied in axis of rotation, angle of rotation, and direction of rotation. We provided the same prompt as in Test 3a.

*Test Using AR with Coordinate System, Angle of Rotation, and Rotation Equation (Test 3c):* We uploaded to AI vision models with 30 screenshots of AR-Classroom where the coordinate system, rotation angle, and corresponding rotation matrix equation are all visible (Figure 7c). We provided the same prompt as in Test 3a and GPT-4 gave very satisfactory results each time we tested, as explained below. Furthermore, even with a simpler prompt "*Explain the rotation process using minimum steps of rotation in terms of axis of rotation, angle of rotation, and direction of rotation*", the results are as accurate as the more detailed prompt in Test 3c.

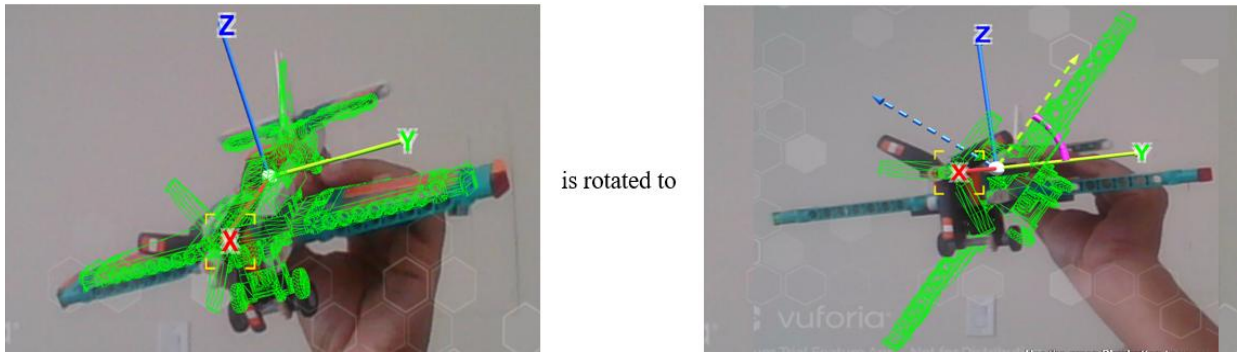

(a)

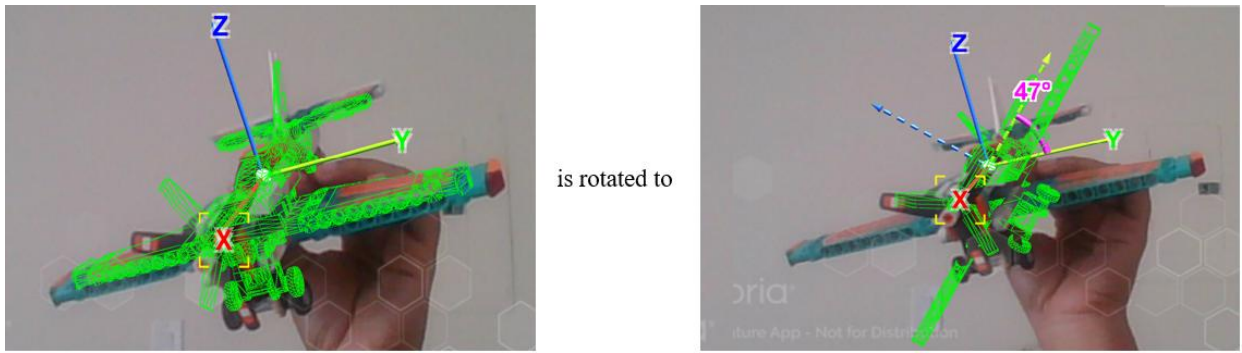

(b)

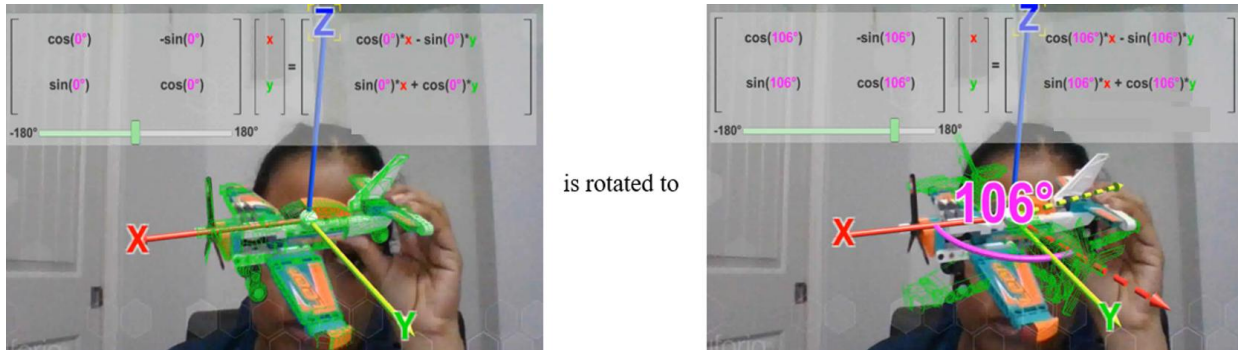

(c)

Figure 7: (a) AR rotation visualization with a coordinate system, (b) AR rotation visualization with a coordinate system and angle, (c) AR rotation visualization with a coordinate system, angle, and a rotation equation

In experiment 3, we tested whether AI vision models can identify axis, angle and direction of rotation individually and then described the whole rotation process combining all three aspects (Figure 8). We tested different versions of AR-Classroom images and observed that the performances of the vision models get better when more descriptive texts are added to the visual input. For example, GPT-4 shows 13.3% accuracy during AR rotation with only the coordinate system (Test 3a), and 63.3% accuracy AR rotation with a coordinate system and an angle of rotation (Test 3b) has. When we provided the whole UI of AR-Classroom (AR with a coordinate system, an angle, and rotation equation) as input, the additional information related to the rotation process helped GPT-4 gain 96.7% accuracy (Test 3c). We can see from Figure 8 that making the angle of rotation and the rotation matrix equation visible in the AR scene each has increased the accuracy significantly.

When we compared the performance of three AI vision models, we GPT-4 could identify text on the image (i.e. angle, axis, and equation) better and recognize the connection between the hand rotation with the corresponding rotation matrix equation (Figure 8). Gemini 1.5 Pro did not perform as well as GPT-4 as it identified the angle of rotation 80% accuracy in test 3b (Figure 8) where angle rotation text was visible on the image that we provided. When the equation was visible in test 3c Gemini 1.5 Pro identified the angle correctly. However, Gemini 1.5 Pro could not identify the axis of rotation from the rotation matrix equation and showed 70.0% accuracy in describing the rotation process. We observed Llama 3.2 identified texts less accurately than the two other AI vision models. It showed less accuracy in identifying angle and axis of rotation and could not recognize the connection between the hand rotation with the corresponding rotation matrix equation properly (30% accuracy).

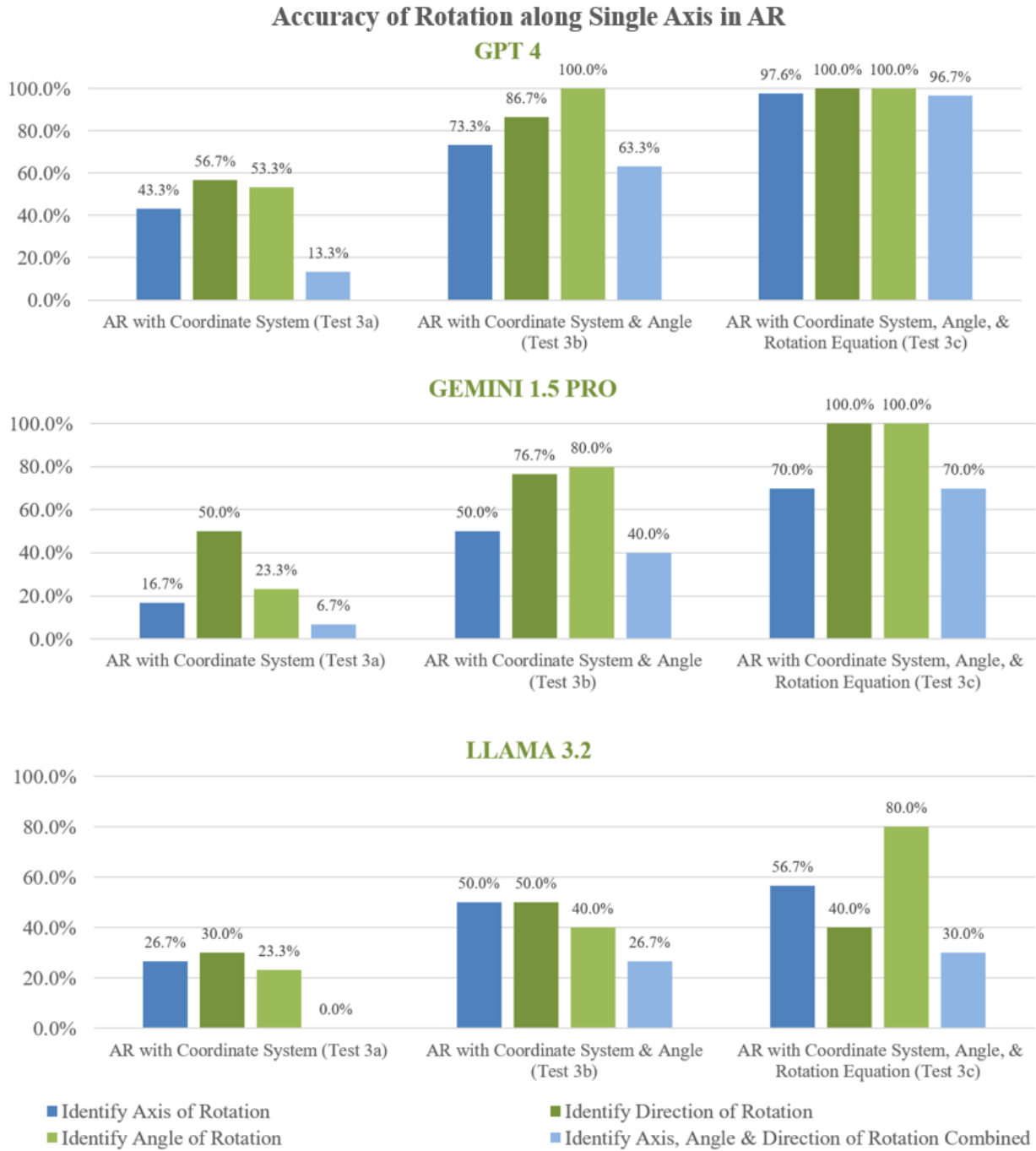

Figure 8: Accuracy of identifying rotation process in Test 3a, Test 3b, and Test 3c

## 5. Discussions

In LLMs, the language model tends to be more accurate for purely text-based tasks, while the vision model is better suited for multimodal tasks involving both images and text [43]. AI vision models integrate image understanding with text, making it powerful for multimodal tasks like interpreting images alongside text descriptions. This model performs well in visual question answering, image-to-text tasks, and recognizing objects in images. However, its accuracy for purely image-based tasks, like object detection or more complex image analysis, may not match specialized vision models and still relies heavily on textual inputs to guide its performance. This supports what we observed in our experiments: three AI vision models gave inadequate performance in the case of Revised PSVT:R (Test 1a) where GPT-4, Gemini 1.5 Pro, and Llama 3.2 showed 16.7%, 23.3%, and 16.7% accuracy respectively. This indicates that it cannot identify complex rotation tasks and cannot identify the similarities between two identical rotation processes with different objects. The cases when it got the right answers could be coincidental.

As a comparison, in 1993, a PSVT:R was administered to 535 first-year engineering students at Michigan Tech during orientation. Approximately 20% of the students (n = 96) scored 60% or less on the PSVT:R, while 80% of the students scored above 60% accuracy [44]. In another study [24] using Revised PSVT:R, the mean score of 1,022 participants was 19.08 out of 30 with a standard deviation of 6.16. Scores ranged from 3 to 30, and 13 participants (1.27%) achieved a perfect score.

Furthermore, providing a 3D coordinate system with the original Revised PSVT:R did not show significant improvement (Test 1b). Using between-subjects ANOVA we found out there is no group difference between three AI vision models' performances ($p$s>.05) for Test 1a and Test 1b. Another between-subjects ANOVA found out there is no significant difference ($p$s>.05) between Test 1a and Test 1b. Which means adding coordinate system with PSVT:R did not help the performances of three AI vision models.

Running between-subjects ANOVA for Test 2 (Experiment 2) showed no significant group difference ($p$s>.05) between three AI vision models in recognizing Step 1 of Revised PSVT:R with coordinate system (a simplified rotation process).

In the AR environment, we provided better context than the previous experiments. Running a between-subject ANOVA at $p$<.05 level between three tests of Experiment 3 (Test 3a, Test 3b, Test 3c) showed significant group difference, $F(2, 6) = 5.25$, $p$=0.047. Three AI vision models could see pre-image before rotation and image after the rotation in the AR environment (Test 3a). By increasing textual input step by step in the AR environment, we observed improved performance by all AI vision models (Figure 8). Just including the angle of rotation with the coordinate system showed significant improvement in result (Test 3b). Providing a rotation matrix equation as additional information enabled all AI vision models to identify the rotation process correctly (Test 3c). Even with a less detailed prompt and same image as Test 3c, it identified the rotation process more accurately. However, running a between-subject ANOVA between three AI vision models for Experiment 3 did not show significant group difference ($p$s>.05). This result indicates no AI vision model showed significantly better accuracy than other model, they all performed better when more information was included in AR environment.

The use of the AR environment is for the convenience of adding accurate rotation information to the testing for AI vision models. A drawing style of PSVT:R may be used in a similar evaluation process. However, the rendering mode (realistic 3D physical model, wireframe virtual model, colors of rotation arc and angle values, etc.) may have affected the understanding of rotations by AI vision models.

Overall, this result points out that AI vision models show limited spatial intelligence in terms of understanding 3D rotations with 2D image inputs. For applications, it is very promising for future development of AR applications in the education and STEM sectors, where AI vision models can guide a user on task performance by taking real-world data and relevant, superimposed textual and graphical data as input. In an AR setting, users can view physical objects, diagrams, or virtual models while AI provides additional information layered over the scene. For example, in a geometry lesson, a 3D model of a shape could be displayed, and AI vision models could guide the student by offering real-time feedback on angles, dimensions, or rotation processes, enabling more effective learning. This combination of visual and textual data allows AI vision models to better understand spatial relationships and provide meaningful insights.

The result also indicates that the current generative AI doesn't perform well regarding spatial intelligence without additional context information (e.g., rotation axis, angle, and equation). The context information can facilitate AI for understanding spatial transformations and can help train AI to improve its spatial intelligence, so that in the future the context information may not be needed any more.

A major limitation of the research is that the sample size (number of each test) is small as this is a very early evaluation of spatial intelligence of AI in terms of 3D rotations.

## 6. Conclusions and Future Work

Spatial knowledge is critical in fields like architecture, computer graphics, civil engineering, and other STEM disciplines. While generative AI-powered tools can assist students in learning, conversational AI agents still struggle with spatial abilities. AR can enhance these systems by overlaying textual information on spatial diagrams, making the data more comprehensible for AI. This paper investigates AI vision model's spatial capabilities, particularly its proficiency in interpreting 3D rotations, as tested through Revised PSVT:R and rotation images from an AR application. We improvised Revised PSVT:R with additional graphical and textual features to assess if AI vision model's performance changes with supplementary information. All AI vision models had difficulty recognizing complex rotation tasks and identifying the similarities when two rotations are essentially the same but for different objects, as tested in the Revised PSVT:R.

In the AR images, where objects are visualized in 3D space, adding supplementary textual and math information significantly improved AI vision models' accuracy. This suggests that while AI vision models perform well in multimodal tasks, it relies heavily on textual cues for better performance. Additional information, as visualized in AR applications, has the potential to enhance the performance of these AI vision models. These results indicate that in an AR setting, users can view physical objects, diagrams, or virtual models' additional information layered over the scene while AI can provide real-time feedback observing the user's interaction process.

The future work will include more thorough testing with an increased sample size and with different AI tools, fine-tuning AI models and enhancing the spatial intelligence of AI, so that AI can understand spatial transformations including rotations, translations, scales, perspectives, etc. based on various styles (e.g., real-world photo or drawing styles) of image or video input of arbitrary 3D physical objects, with or without context information of axes, text, math equations, etc.

## Acknowledgement

This work is partially supported by National Science Foundation (NSF) Grant #2119549. The authors appreciate the support of the Texas A&M University's team of the NSF RETTL project: Using Augmented Reality and Artificial Intelligence to Improve Teaching and Learning Spatial Transformations in STEM Disciplines, for the AR-Classroom app and discussions on mental rotations and AI.

# Appendices

Some conversation examples from our experiments are included here.

*A. GPT-4 Test Using Revised PSVT:R*

> **Sample question from authors (June, 2024):**
>
> ```
> [Revised PSVT:R Question 2 screenshot image upload to GPT-4]
> ```
>
> **Sample answer from GPT-4, whose answer is B in the end (not necessarily to be correct):**
>
> ```
> To answer question 2, we need to visualize the rotation applied to the
> object in the middle line and find the matching option from the bottom line.
>
>
>
> **Step-by-Step Analysis:**
>
> 1. **Top Line**: Shows an initial object and its rotated version.
>
> 2. **Middle Line**: Shows another object labeled "As."
>
> 3. **Bottom Line**: Provides five possible rotated versions labeled A, B, C,
> D, and E.
>
>
>
> **Top Line Rotation:**
>
> - The initial object in the top line is rotated such that the front face
> (with the cutout) moves to the left, and the side face is on the top.
>
>
>
> **Applying the Same Rotation:**
>
> - The object labeled "As" in the middle line will undergo the same rotation.
>
> - Visualize the rotation where the curved surface, currently facing forward,
> will move to the left, and the flat surface will face upward.
>
>
>
> **Matching the Rotated Object:**
>
> - Option A does not match the rotation.
>
> - **Option B** matches the rotation.
> ```

```
- Option C does not match the rotation.

- Option D does not match the rotation.

- Option E does not match the rotation.

Thus, the correct answer is **B**.
```

B. Test Using Revised PSVT:R with a coordinate system axes

**Sample prompt by authors into GPT-4 with the uploaded image based on Revised PSVT:R with axes added (September, 2024):**
The first line shows an object and how it is rotated. The task is to study how the object in the top line of the question is rotated. Picture in your mind what the object shown in the middle line of the question looks like when rotated in exactly the same manner. Select from among the five drawings (A, B, C, D, or E) given in the bottom line of the question the one that looks like the object rotated in the correct position.
Objects in the Example: The top line shows the initial object and its rotated version. The middle line shows another object labeled "As". The bottom line provides five possible rotated versions labeled A, B, C, D, and E. To solve the example, you would visualize the rotation applied to the middle object ("As") and choose the correct rotated version from the options provided.

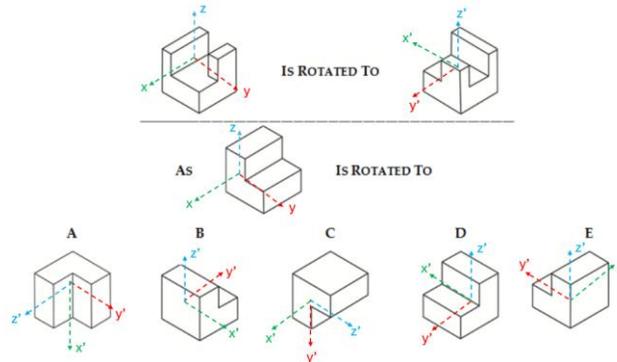

**Sample answer from GPT-4:**
To solve this problem, let's walk through the steps in detail by analyzing the rotation transformation applied to the top object and then applying it to the "As" object.
1. Analyze the Rotation in the Top Row: In the top row, we see that the initial object is rotated to achieve a new orientation. Observing the axes:
- The x-axis (green) in the initial object is rotated to the z-axis position (upward).
- The z-axis (blue) rotates to the x-axis position (forward).
- The y-axis (red) stays in the same orientation.
2. Apply the Same Rotation to the "As" Object: Now, we apply this rotation to the middle object labeled "As":
- Rotate the x-axis (green) of "As" to where the z-axis would point (upward).
- Rotate the z-axis (blue) of "As" to the position of the x-axis (forward).
- Keep the y-axis (red) in the same orientation.
3. Identify the Matching Option:

- Now, look at each option (A, B, C, D, E) in the bottom row to find the one that matches the rotated "As" object.
- Option C has the correct orientation of the axes:

The green x-axis points upward.
- The blue z-axis points forward.
- The red y-axis is unchanged.

Therefore, the correct answer is Option C.

*C. Test of Step 1 of Revised PSVT:R with a coordinate system*

**Sample prompt by authors into GPT-4 with the uploaded image based on Step 1 of a Revised PSVT:R question with axes added (September, 2024):**
Explain the rotation process using minimum steps of rotation in terms of axis of rotation, angle of rotation, and direction of rotation.

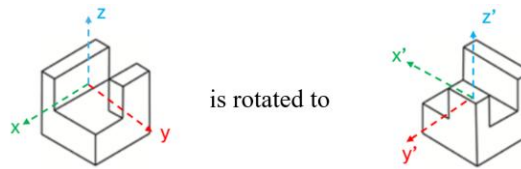

is rotated to

**Sample answer from GPT-4:**
To describe the rotation process shown in this image with minimal steps, let's focus on the changes in axis orientation between the initial and rotated object.
Step-by-Step Rotation Process
1. Identify the Rotation Needed:
- In the initial object, the x-axis (green) is pointing to the left, the y-axis (red) is pointing forward, and the z-axis (blue) is pointing upward.
- In the rotated object, the x'-axis (green) is pointing forward, the y'-axis (red) is pointing to the right, and the z'-axis (blue) remains pointing upward.

This indicates a rotation that affects the x and y axes while keeping the z-axis unchanged.
2. Rotation Details:
- Axis of Rotation: Z-axis.
- Angle of Rotation: 90 degrees.
- Direction of Rotation: Clockwise when viewed from above.

Explanation
By applying a 90-degree clockwise rotation around the z-axis, the x-axis moves from pointing left to pointing forward, and the y-axis moves from pointing forward to pointing right, while the z-axis remains in its original upward orientation.
This minimal rotation successfully transforms the initial object to match the rotated version shown.

*D. Test of AR Rotation with Coordinate, Angle of Rotation, Corresponding Matrix Equation*

**Sample prompt by authors into GPT-4 with the uploaded image based on AR-Classroom (October, 2024):**
Explain the rotation process with axis of rotation, angle of rotation and direction. Note: All red lines represent X-axis, all yellow-green lines represent Y-axis, all blue lines represent Z-axis. The dotted line is the orientation of axis after rotation. The axis of rotation does not have a dotted counterpart of the same color. Follow right hand rule for the direction of rotation.

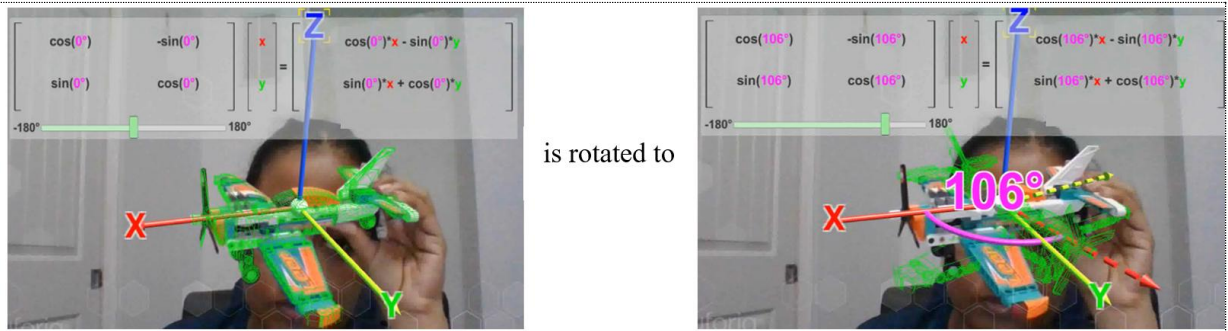

**Sample Answer from GPT-4:**
In this image, we see a rotation of the model from an initial position (left) to a rotated position (right) around the **Z-axis** (blue line) by an angle of **106°**.
Here's a detailed breakdown of the rotation process:
1. **Axis of Rotation**:
   The **Z-axis** (blue line) serves as the axis around which the model rotates. This axis remains fixed and does not have a dotted counterpart because it acts as the pivot for the rotation.
2. **Angle of Rotation**:
   The rotation angle is **106°**, as indicated in purple. This positive angle represents a counterclockwise rotation around the Z-axis.
3. **Direction of Rotation**:
   Using the **right-hand rule**, if you point your right thumb along the Z-axis, your fingers curl in the direction of the rotation. Since the angle is positive, the rotation proceeds counterclockwise when viewed from the positive direction of the Z-axis.
4. **Resulting Orientation**:
   After rotating 106° around the Z-axis, the **X-axis** (red line) and **Y-axis** (yellow-green line) shift orientation. The dotted red and yellow-green lines in the right image show the new positions of the X and Y axes, respectively, after the rotation. The X-axis rotates to a new position indicated by the dotted red line, and the Y-axis shifts to the dotted yellow-green line, showing how the model's alignment changes in 3D space.

In summary, this transformation represents a **106° rotation around the Z-axis**. The Z-axis remains as the fixed axis, while the X and Y axes change orientation, rotating counterclockwise according to the right-hand rule for a positive angle.